\documentclass{ruthesis}
\usepackage{graphicx}
\usepackage{subcaption}
\usepackage[hidelinks]{hyperref}
\begin{document}

\title{Face Identification and Clustering}
\author{Atul Dhingra}
\program{Computer Science}
\director{Dr. Vishal Patel, Dr. Ahmed Elgammal}
\approvals{3}
\submissionyear{2017}
\submissionmonth{May}

\abstract{

In this thesis, we study two problems based on clustering algorithms. In the first problem, we study the role of visual attributes using an agglomerative clustering algorithm to whittle down the search area where the number of classes is high to improve the performance of clustering. We observe that as we add more attributes, the clustering performance increases overall. In the second problem, we study the role of clustering in aggregating templates in a 1:N open set protocol using multi-shot video as a probe. We observe that by increasing the number of clusters, the performance increases with respect to the baseline and reaches a peak, after which increasing the number of clusters causes the performance to degrade. Experiments are conducted using recently introduced unconstrained IARPA Janus IJB-A, CS2, and CS3 face recognition datasets.

}

\beforepreface

\acknowledgements{

I would like to first thank my thesis advisor, Dr. Vishal Patel who provided gave me inspiration and encouragement throughout. I would also like to thank my thesis co-advisor, Dr. Ahmed Elgammal who helped me navigate through this journey. I would also like to thank all my lab-mates for the exchange of ideas, academic and otherwise. Finally, I express my profound gratitude for my family who has helped me arrive at this point in my academic career. \\

}
\dedication{
I dedicate this thesis to my family.
}
\afterpreface

\chapter{Face Recognition}
\section{Introduction}
\begin{figure}[h]

\begin{center}
\includegraphics[width=12cm]{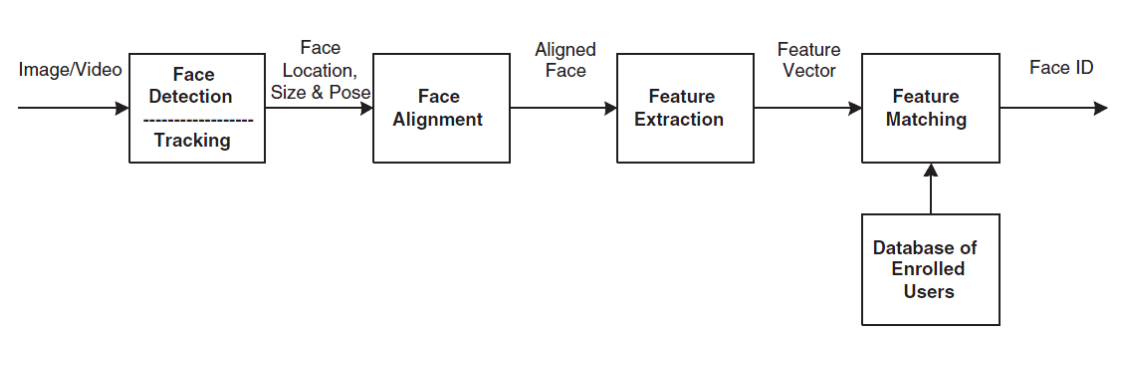}
\caption{Face Recognition Pipeline}
\label{fig:FR_pipeline}
\end{center}
\end{figure}

Face recognition has been actively studied over the past few decades which has led to satisfactory performances in recognition rates in controlled scenarios. But, in an unconstrained environment, face recognition is still a hard problem. A number of datasets have been thus developed to study face recognition in these scenarios that include LFW \cite{LFW}, PubFig\cite{pubfig} and IJBA\cite{IJBA}. The intuitive pipeline\cite{HandbookFR} is shown in  figure \ref{fig:FR_pipeline} for face recognition, that includes face detection and tracking, face alignment, feature extraction and matching, described in sections below.

\subsection{Acquisition} 
There are a few challenges that hinder the progress of face recognition in an unconstrained environment, which include challenges such as pose, illumination, and expression (PIE). Some of the other notable challenges include aging, cosmetics and resolution of the image. A lot of datasets have been developed that provide challenging media(images, videos, templates) so that algorithms can be developed to deal with these issues. Yale and YaleB\cite{YaleB} was introduced in 1997 that highlighted the challenges in illumination conditions, AR dataset\cite{AR} in 1998 highlighted occlusion apart from different emotions, and illuminations. Some of the more notable datasets in recent past are LFW\cite{LFW} and PubFig\cite{pubfig}	that contain huge amounts of images and deal with the face representation in the wild. One of the most challenging datasets as of now is the recently introduced unconstrained IARPA Janus IJB-A, CS2 and CS3 face recognition datasets\cite{IJBA}.
\subsection{Normalization and Alignment}
Some of the pose and illumination artifacts are removed by normalization. There has been a lot of work that deals with this task. Depending on the applications, the issue of normalization can either be handled during the acquisition phase, where during the collection of the database, the acquisition parameters, such as capture device, ambient light are fixed. But, in the case where we want to develop algorithms invariant to these artifacts in unconstrained settings, learning from data in such preferential environment is averse to learning in the real-world settings. In such a case, post-processing of the collected data is done. 
\subsubsection{Illumination Normalization}
We can handle illumination normalization during the acquisition phase, by making sure that the illumination remains the same throughout. As some of the datasets are collected in the real world settings, the natural illumination affects the final dataset. In such a case active devices such as thermal infra-red images, near infra-red images etc. can be used that provides its own light source to illuminate the object. \\
In case this is not possible, such as images in the wild, normalization is done during post-processing to generate illumination invariant features. This can be done by using methods such as linear subspace, illumination cone, generalized photometric stereo, photometric normalization, reflectance model \cite{Illumination} etc. Some of the most studied models include, Self Quotient Images \cite{SQI},Logarithmic Total Variation \cite{LTV}, Gradient Faces\cite{GF}, Robust Albedo Estimation\cite{RAE}.
\subsubsection{Pose Normalization}
The images captured during the data acquisition phase can be constrained such that the pose of the captured images are consistent. But, is not the best solution, as even a slight error in capturing would result in a completely different image vector. Therefore, in such a case pose normalization is done during post-processing. The approach, in this case, is to find landmarks in the image that would remain consistent throughout, no matter how much shifted the image is. Some of these landmarks include the eyes, the nose, and the lips. Once these landmarks are detected, the image can be normalized based on these set of points. One such method that takes into account such an approach is called Geometric warping\cite{GW} where in-place pose normalization can be achieved. But, this approach cannot help in the case where there is an out-of plane rotation, for a case more robust methods are required. This follows from the fact that in an out-of plane rotation, pitch, roll, and yaw all have to be normalized. Some of the more used methods studied are, Incremental face alignment\cite{Asthana}, Deep Face Alignment\cite{DFA},Face Frontalization\cite{Hassner}.
\subsection{Features/Recognition}
Features are distinct and unique properties of an entity, that can be used to distinguish it from others. These features are important as they form the framework for recognition of these entities. In a face recognition system, facial features could include, the shape of a person's face, eye color, the distance between eyes, etc. These features could either be hand-crafted, or they could be learned features. 
\subsubsection{Hand-Crafted features} 
Hand-crafted features as the name suggests is created manually by observing uniqueness in some aspect of the object. At the lowest levels, edges, lines, and corners form features, in a complex object, such as a face, a combination of these low-level features by hand is known as hand-crafted features. There are a few hand-crafted features that have been used extensively, such as SIFT\cite{SIFT}, HOG\cite{HOG}, LBP\cite{LBP}, etc. In such a framework, a classifier is trained using the hand-crafted features.The classification/recognition can be done using  SVM\cite{SVM} , SRC\cite{SRC} and Subspace methods such as PCA\cite{PCA}, LDA \cite{LDA} etc.

\subsubsection{Learned Features}
Instead of coding the features by hand, features can also be learned from the data. This ensures an optimal representation given the data. At the end, a simple classifier can be used for classification. There are a few methods that are used in such a scenario, which include Dictionary Learning, Neural Networks etc.
\section{Protocols}\label{frprotocols}
Recognition is a term with wide scope when it comes to Face biometrics, as it encompasses a lot of authentication protocols, there are a few widely used authentication types that have been described below. 
\subsection{Identification}
In an identification problem, the question asked is, whether a given person exists in our system or not. The output from such a system is either Identified or Not-identified depending on whether that person exists in the given database.
\subsection{Verification}
In a verification protocol, given an instance of a user, we check if it matches the sample of the same user in our system. The output in such a scenario is a similarity score which defines how closely the new sample matches to the one already in the system
\subsection{Search}
In a search scenario, given a query image, we need to find all the instances of that person in the database. The output, in this case, top-k hits of the subject

\section{Metrics}
\subsection{Error Statistics}
A few of the more commonly used error statistics are False Match(Type I Error), False Non-Match (Type II Error), True-positive Identification Rate(TPIR),  False-positive Identification-error Rate (FPIR).
\subparagraph{True-positive Identification Rate(TPIR)}
The True-positive Identification Rate (TPIR) is the proportion of identifications by enrolled subjects in which the subject’s correct class is returned. \cite{BiometricsReport}
\subparagraph{ False-positive Identification-error Rate (FPIR)}
The False-positive Identification-error Rate (FPIR) is the proportion of identifications by users not present in the system, which is returned. FPIR cannot be computed in closed-set identification, as all users are enrolled in the system
\cite{BiometricsReport}
\subsection{Decisions}
\subsection{Metric curves} 
\begin{figure}[h!]
\begin{center}
\includegraphics[width=10cm, height=6cm]{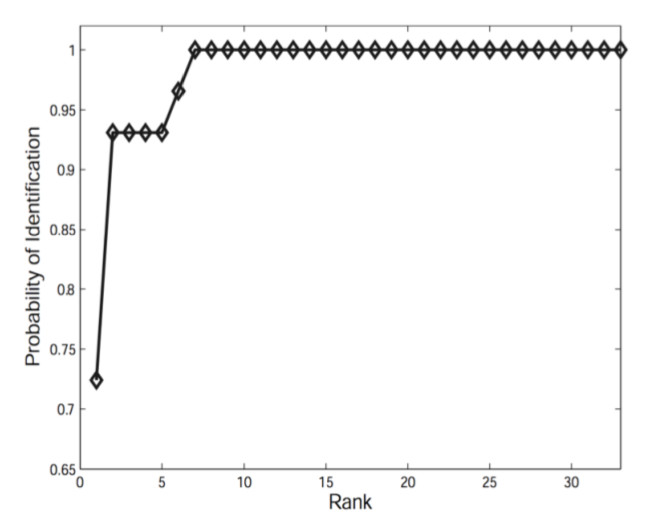}
\caption{CMC Curve\cite{HandbookFR}}
\label{CMC}
\end{center}
\end{figure}

There are a few metric curves that are used to plot the decisions, that include Reciever Operating Characteristics(ROC), Detection Error Tradeoff(DET), Cumulative Match Curve(CMC).
\subparagraph{CMC}
A CMC curve plots the Probability of identification versus the Rank as shown in figure \ref{CMC}

\subsection{Result interpretation}
The result interpretation depends on the type of face recognition application. Some of the more used interpretations include Accuracy, Precision and Recall and F-Measure.
\subparagraph{F-measure}
The F-measure is given in equation \ref{eq:F-beta} where P is Precision and R is Recall
\begin{equation}
\label{eq:F-beta}
F_{\beta} = \frac{(\beta^2+1)P.R}{\beta^2P + R}
\end{equation}
The F-1 measure is widely used where $\beta=1$, such that F-1 measure is the harmonic mean between precision and recall. The value of F-measure, therefore, is always between 0 and 1, and the higher the value, better is the performance of the recognition algorithm.
\subparagraph{Precision and Recall}
Precision is defined as the ratio of True positives(TP) to the sum of True positives and False Negatives(FN) as shown in figure \ref{fig:PR}	
\begin{equation}
\label{fig:RE}
Precision=\frac{TP}{TP+FP}
\end{equation}

Recall is similarly defined as the ratio of true positives over the sum of true positives and false positives(FP) as in figure \ref{fig:RE}
\begin{equation}
\label{fig:PR}
Recall=\frac{TP}{TP+FN}
\end{equation}
\chapter{Face Clustering}
\section{Introduction}
\begin{figure}[h!]
\begin{center}
\includegraphics[width=10cm, height=2cm]{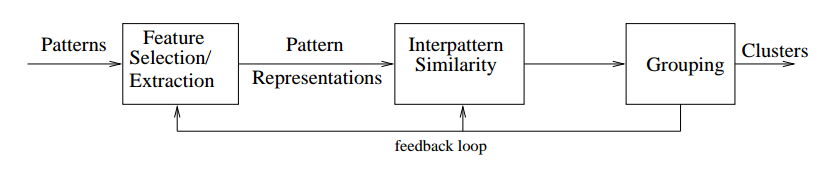}
\caption{Pipeline for Clustering}
\label{fig:ClusteringPipeline}
\end{center}
\end{figure}

Clustering is an unsupervised classification of patterns such as data items, feature vectors, or observations. In such a setting, given unlabelled data points, we have to group them based on a metric($\ell_2$,$\ell_p$, Mahalanobis etc.). Clustering is a difficult problem, as we need to know a priori about the number of clusters or the stopping criterion. Clustering has a lot of applications such as exploration, segmentation in cases where the prior information about the data is not available. The pipeline\cite{Survey_Clustering} for clustering is given in figure \ref{fig:ClusteringPipeline}

A good representation the given data points/patterns is achieved by feature extraction. Once these features are computed, the clusters are merged/divided based on the inter-pattern similarity and the type of clustering. This process goes on until a stopping criterion such as a distance threshold or max number of clusters is met.

\section{Clustering Techniques}
\subsection{Hierarchical}
Hierarchical clustering seeks to build a hierarchy of clusters such that it yields a dendrogram that represents the nested grouping of patterns and similarity levels\cite{Survey_Clustering}. These fall into two categories, agglomerative clustering, and divisive clustering.
\subsubsection{Agglomerative} \label{Agglomerative}
This is a bottom-up approach where each observation starts as an independent cluster, and pairs of clusters are merged based on the hierarchy and a stopping criterion. The merging of the clusters is based on certain linkage criterion, such as Single Link where the minimum distance between the points is used to merge the cluster. In the case of complete-link clustering, the clusters are merged based on the maximum distance between the data points of the two clusters. There are other order statistics that are used such as mean, centroid, group average, etc. to perform these linkages as well. 
\subsubsection{Divisive} In a divisive clustering framework, a top-down approach is followed such that all the data points start out in a single cluster,  and they are split into different clusters moving down the hierarchy.
\subsection{Partitional}
In the case where construction of dendrograms is computationally inefficient/impossible, partitional methods are employed where a single partition of the data is obtained instead of a structure. The issue with using partitional clustering techniques is the fact that we need to know a priori the number of clusters/ partitions we need to perform. Partitional clustering is produced by optimizing a criterion function defined either locally or globally\cite{Survey_Clustering}. Some of the most common criterion used are squared error method as represented in equation \ref{eq:SquaredError} \cite{Survey_Clustering}, where  X is the patterns set of the clustering L, which  contains K clusters, such that $x_i^{(j)}$ is the $i^{th}$ pattern belonging to the $j^{th}$ cluster and $c_j$ is the centroid of the $j^{th}$ cluster

\begin{equation}
\label{eq:SquaredError}
e^2(X,L)=\sum_{j=1}^K \sum_{i=1}^{n_j} ||x_i^{(j)}-c_j||^2
\end{equation}

A widely used method that uses squared error criterion is the k-means algorithm, where k points are randomly picked as the centroid and the cluster's center are recomputed until convergence by assigning each point to the cluster with the closest centroid.
\section{Evaluation}\label{Pairwise F-1}
The ultimate aim for clustering algorithm is to attain high intra-cluster similarity and low inter-cluster similarity. There are few evaluation metrics that are widely used to access the quality of the clustering. Some of these are Purity, Precision, and Recall, F-measure and compactness\cite{Cluster_eval}. In our work, we use pair-wise precision and recall as defined in \cite{Millions}\\
\textit{Pairwise Precision} is the same class fraction of pairs of data points within a cluster over the total number of same cluster pairs within the dataset. \cite{Millions}. \\
\textit{Pairwise Recall} is the fraction of within class pairs of data points, that are placed in the same cluster, over the total number of same-class pairs in different clusters. \cite{Millions}. 

\section{Recent Works and Motivation}
\begin{figure}[h!]
\begin{center}
\includegraphics[width=10cm, height=4cm]{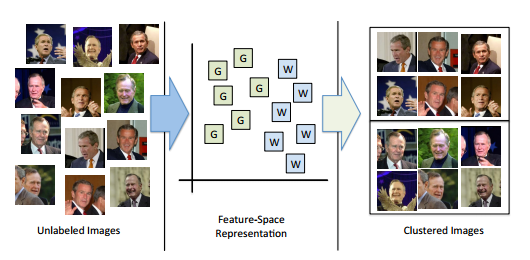}
\caption{Pipeline for Face Clustering \cite{Millions}}
\label{fig:FC}
\end{center}
\end{figure}

In the problem of clustering faces, given unlabelled face images, we need to divide them into clusters, using a good feature space representation and a distance metric as shown in figure \ref{fig:FC}\cite{Millions}.

There has been a lot of work in the area of face clustering that tries to improve the clustering accuracy. Zhu et al.\cite{Rank-Order} came up with Rank-Order Distance that is robust to both noise and outliers and can handle non-uniform cluster distribution like varying densities, shapes, and sizes of clusters. It calculates the dissimilarity between two faces based on their neighbouring information using $\ell_1$ distance motivated by the fact that the same person shares top neighbours.  The sub-clusters formed due to variation in pose illumination and expression, are subsequently merged agglomeratively using rank-order distance using a certain threshold or cluster level rank order distance to avoid the problem of too many high-precision, tight sub-clusters in the case of just using rank-order distance. \\
Otto et al\cite{Millions} used the same idea as Zhu et al. \cite{Rank-Order} on a larger scale, and therefore modified the algorithm to work on a large data setting. The effective and efficient Rank-order clustering algorithm used k-d tree algorithms to compute a small list of nearest neighbour, as the input size of data in order of millions, generating all the neighbours, as in the case of Zhu et al. \cite{Rank-Order} would be computationally hard. It used a single linkage agglomerative clustering algorithm based on a threshold to further compute the clusters and uses a pairwise F-measure to report the results on LFW dataset\cite{LFW}. \\
Zhu et al\cite{Context-Assisted} came up with an algorithm to iteratively merge high precision clusters based on heterogeneous context information such as common-scene, people co-occurrence, human attributes and clothing information,  such that the resulting clusters also have high recall. 

\begin{figure}[h!]
\centering
\includegraphics[width=0.5\textwidth]{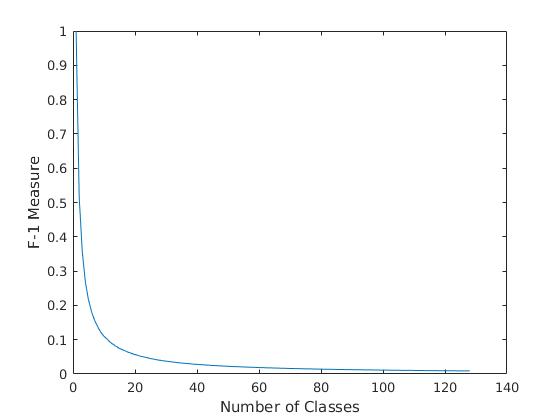}
\caption{F-measure plot as the number of classes increases}
\label{fig:Clustering Hard}
\end{figure}

Clustering is hard as the performance decreases as the number of classes increases as it is evident in figure \ref{fig:Clustering Hard}. Therefore our work is motivated by this challenge to whittle down the search domain in clustering using visual attributes to improve the clustering accuracy.

\section{Experiment}
\begin{figure}[h!]
\begin{center}
\includegraphics[width=10cm, height=4cm]{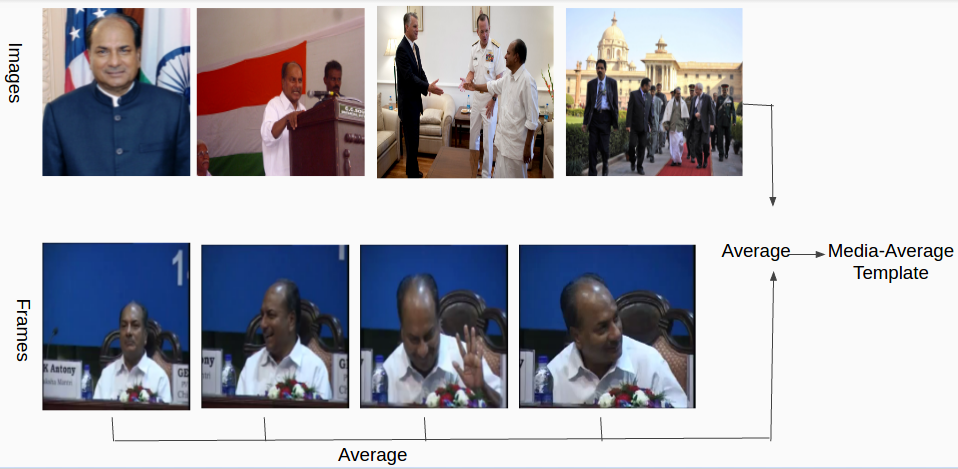}
\caption{Media Average template \cite{IJBA}}
\label{fig:Media Averaging}
\end{center}
\end{figure}

In our work, media averaging is done on CNN features that are computed from the IJBA CS2\cite{IJBA} samples in order to obtain templates. As the CS2\cite{IJBA} follows a template to template matching protocols, we perform clustering on these media averaged templates. Media averaging is shown in figure \ref{fig:Media Averaging}, where the video frames with the same media ID are averaged, and the resultant is then averaged with the images that belong to the same template ID. 

Once these templates are obtained, we use agglomerative clustering as defined in section \ref{Agglomerative}	where each template starts out as a different cluster are merged based on the stopping criterion of max number of clusters, as we have prior information of classes from the dataset. We use the average linkage with the cosine metric for merging these clusters based on the inter-pattern similarity. The templates are further divided into disjoint sets based on ground truth attributes from CS2 \cite{IJBA}. The template is divided into a Male subset, a female subset. The male subset is further divided into two different disjoint subsets based on the skin color attribute.The accuracy of the algorithm is reported based on pairwise F-1 score described in section \ref{Pairwise F-1}
\section{Results}
The algorithm is evaluated on IJBA CS2 dataset \cite{IJBA} that contains 500 subjects with 5,397 images and 2,042 videos split into 20,412 frames. The IJBA CS2 evaluation protocol consists of 10 random splits that contain 167 gallery templates and 1763 probe templates. The algorithm is evaluated on these 10 splits on JC's\cite{JC} and Swami's\cite{Swami} deep features. The evaluated results on Swami's \cite{Swami} features are shown in table \ref{table:SwamiCluster} and figure \ref{fig:Swami_cluster}. The evaluated results on JC's\cite{JC} features are shown in table \ref{table:JC_cluster} and figure \ref{fig:JC_cluster}
\begin{table}[h!]

\centering

\begin{tabular}{|l|l|l|l|l|l|}
\hline
Split   & Base             & Male             & Female           & M+Skin 1         & M+Skin 3         \\ \hline
1       & 0.7281           & 0.7349           & 0.7542           & 0.7417           & 0.82             \\ \hline
2       & 0.7134           & 0.6756           & 0.8364           & 0.7384           & 0.7728           \\ \hline
3       & 0.6817           & 0.7025           & 0.7001           & 0.74             & 0.7336           \\ \hline
4       & 0.7349           & 0.7309           & 0.7676           & 0.7633           & 0.7683           \\ \hline
5       & 0.6066           & 0.6133           & 0.6418           & 0.6517           & 0.648            \\ \hline
6       & 0.6756           & 0.6729           & 0.7577           & 0.7213           & 0.7524           \\ \hline
7       & 0.7309           & 0.7651           & 0.7294           & 0.7513           & 0.816            \\ \hline
8       & 0.6561           & 0.6875           & 0.616            & 0.7648           & 0.8001           \\ \hline
9       & 0.6531           & 0.6845           & 0.7939           & 0.7391           & 0.7095           \\ \hline
10      & 0.6645           & 0.6907           & 0.6875           & 0.7296           & 0.7504           \\ \hline
Average & \textbf{0.68449} & \textbf{0.69579} & \textbf{0.72846} & \textbf{0.73412} & \textbf{0.75711} \\ \hline

\end{tabular}
\caption{Evaluation of algorithm on Swami Features\cite{Swami}}
\label{table:SwamiCluster}
\end{table}

\begin{figure}[h!]
\centering
\includegraphics[width=0.5\textwidth]{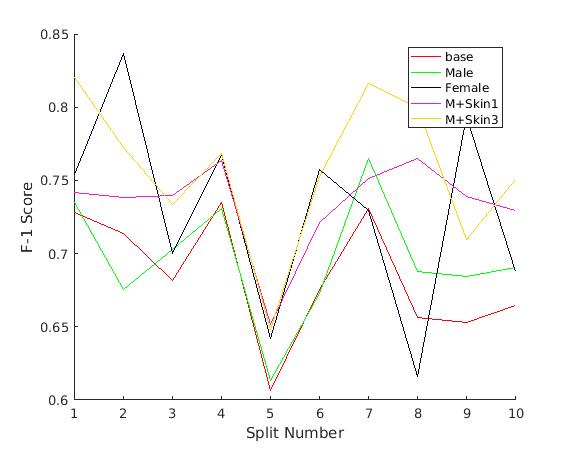}
\caption{F-1 score of the experiment using Swami\cite{Swami} features}
\end{figure}

\begin{table}[h!]
\centering
\begin{tabular}{|l|l|l|l|l|l|}
\hline
Split   & Base             & Male             & Female           & M+Skin 1         & M+Skin 3         \\ \hline
1       & 0.6847           & 0.6854           & 0.7591           & 0.6807           & 0.7424           \\ \hline
2       & 0.682            & 0.6512           & 0.753            & 0.6334           & 0.728            \\ \hline
3       & 0.6356           & 0.6675           & 0.6673           & 0.7044           & 0.6997           \\ \hline
4       & 0.6716           & 0.6597           & 0.7197           & 0.7111           & 0.7325           \\ \hline
5       & 0.5658           & 0.5706           & 0.6036           & 0.5999           & 0.6083           \\ \hline
6       & 0.6633           & 0.6385           & 0.7586           & 0.651            & 0.738            \\ \hline
7       & 0.6832           & 0.6931           & 0.6844           & 0.7266           & 0.8204           \\ \hline
8       & 0.6534           & 0.706            & 0.5862           & 0.7242           & 0.7833           \\ \hline
9       & 0.6157           & 0.6563           & 0.6405           & 0.7109           & 0.6677           \\ \hline
10      & 0.6663           & 0.6783           & 0.6712           & 0.7133           & 0.7694           \\ \hline
Average & \textbf{0.65216} & \textbf{0.66066} & \textbf{0.68436} & \textbf{0.68555} & \textbf{0.72897} \\ \hline

\end{tabular}
\caption{Evaluation of algorithm on JC Features\cite{JC}}
\label{table:JC_cluster}
\end{table}

\begin{figure}[h!]
\centering
\includegraphics[width=0.5\textwidth]{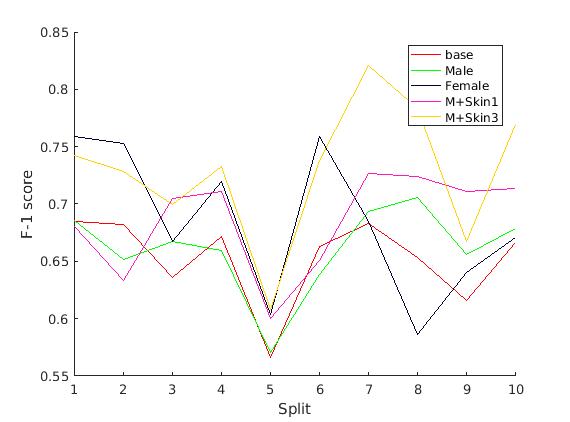}
\caption{F-1 score of the experiment using JC\cite{JC} features}
\label{fig:JC_cluster}
\label{fig:Swami_cluster}

\end{figure}

\section{Conclusion}
We observe  in table \ref{table:SwamiCluster} and table \ref{table:JC_cluster} that as we use more attributes, the clustering result improves. We can, therefore, assert that by using visual attributes we are narrowing down the search domain of the algorithm to boost the performance of clustering.

\chapter{Video Based face tracking and Identification}
\section{Introduction}
In this work, we focus on a face identification task where the target is a multi-shot video and is annotated only once in one of the frames, and we need to search the annotated subject in a given gallery of images. 
The advantage over image to image retrieval in this case is that with a probe video, we have a lot more information and exemplars of the subject of interest and we can leverage this information to come up with a more robust representation that is invariant to the PIE challenges in face recognition. \\
Traditionally, for a video to image retrieval task, the probe video is single shot where frame by frame bounding box of the subject of interest is provided as in the case of Youtube Faces \cite{YTF}. For our work, we study an open set 1:N protocol using full motion video as probe where the probe video is multi-shot. In this setting, the subject of interest is annotated only for one of the frames, and the subject may or may not reappear in the subsequent shots. Therefore, matching a subject of interest from multi-shot video to gallery is a difficult task as we cannot use the traditional methods of a frame by frame bounding box tracking for the target face, because tracking algorithms are prone to drifting.\\
A baseline approach to this problem is just to use the initial representation of the user annotated face of the subject to search for the subject in the gallery. But, the initial representation may not always be full frontal and devoid of any pose, illumination and expression variations. Hence, finding the subsequent appearance of the subject in the video is required to come up with a very robust representation of the subject. This is relatively easy in a single-shot video, where the entire video is a single shot, and there is no break in continuity. This can be achieved by making use of the temporal information and tracking the subject throughout in the video. But, in the case of multi-shot video, this task is relatively hard in a multi-shot video. 
\section{Motivation and Recent Works}
The problem of face recognition as described in section \ref{frprotocols}, can be looked at in the terms of face verification and face identification. In face verification protocol one-to-one similarity is computed between the probe and the reference image. In face identification, on the other hand, one-to-many similarity between the probe and gallery is computed. With LFW\cite{LFW} the face there were attempts to solve the face identification in the case where the dataset was unconstrained. Even so, there was a near-frontal selection bias while constructing the LFW\cite{Template}, hence the results are not representative of the set containing large in-the-wild pose variation. Also, because recent studies, \cite{Human Performance} suggest the algorithmic performance of Face recognition algorithms is sub-par to humans, performance on unconstrained datasets with extreme pose, illumination, and expression are still lagging. One such challenging dataset is IJBA\cite{IJBA} that provides protocols for template-based verification and identification. The dataset consists of images and videos of subjects that are manually annotated and the performance evaluation is over a template, such that set of all media is combined into a single representation. Generating a robust representation in the form  of a template is of utmost importance due to the large variation in pose, illumination, and expression. In our work, we improve an existing algorithm by template aggregation using clustering. \\

There has been some work on templates and multi-shot video to gallery retrieval that has motivated our work in this direction. N. Crosswhite et al. \cite{Template} presented template adaptation, a type of transfer learning that works on the IJBA dataset \cite{IJBA} on one-to-many face identification protocol using CNN features,
and a template specific one-vs-rest linear SVM. In their work, they learned a transfer learning mapping such that the source domain is the CNN features learned, and the target domain is the template of new subjects. This work uses encoding from the penultimate layer of VGG-Face\cite{VGG-Face} using an anisotropically scaled face crop of 224x223x3, followed by learning an L2-regularized L2-loss primal SVM with class weighted hinge loss objective\cite{Template} as expressed in equation \ref{eq:Template}.

\begin{equation}
min_w \frac{1}{2} w^Tw +C_p \sum_{i=1}^{N_p}max[0,1-y_iw^Tx_i]^2 +
C_n \sum_{j=1}^{N_n}max[0,1-y_jw^Tx_j]^2
\label{eq:Template}
\end{equation}

such that $C_p$ is the regularization constant for $N_p$ positive observations obtained via average media encodings in the template, and $C_n$  for negative observation obtained via large external negative features. \\
Ching Hui et al. \cite{TFA} combine the work of Template Adaptation \cite{Template} and context-assisted clustering \cite{Context-Assisted} to propose a Target Face Association(TFA) technique \cite{TFA} that retrieves a set of representative face images from multi-shot video that is likely to have the same identity as the target face which is then used as a robust representation based on which the subject is looked up in a gallery of images.  An OTS tracking technique\cite{TFA-OTS} is used to track the target face. These images are treated as the initial positive training set($S_p$). The faces are pre-associated \cite{TFA} by selecting highest Intersection over Union(IOU)\cite{TFA-IOU} of face detection bounding box with the  with tracking bounding boxes for the first k-frames. Ching Hui et al. learns a target specific linear SVM iteratively from pre-associated face images(positive samples) from the target video and negative samples($S_n$) obtained from the cannot-link constraint\cite{Context-Assisted}. In the cases where the target video cannot establish cannot-link constraints, due non-existence, an external dataset ($S_b$) is prepared for negative instances of the SVM.  Their work uses two different models, wherein model one, the linear SVM is solved using the max-margin framework, where the training data is the union of all the three sets, i.e $\{(x_i,y_i)|i \in (S_p \cup S_n \cup S_b)\}$ are used for training.  
In model 2, the set $S_b$ is used only when there are no within-video negative instances. 
The robust face representation\cite{TFA} is given in equation \ref{eq:TFA_Avg} , 
\begin{equation}
x^{fa}=\frac{1}{|A|}\sum_{i \in A}x_i
\label{eq:TFA_Avg}
\end{equation}

\section{Method}
Once the TFA \cite{TFA} algorithm outputs the positive samples from the SVM, it simply averages these features as shown in equation \ref{eq:TFA_Avg} to obtain the robust representation. In our work, however (TFA-C), we leverage a clustering algorithm to aggregate the features at the end of TFA. We use an Approximate Nearest Neighbour k-means++ using VLFeat library \cite{vlfeat} algorithm  such that the k data points that are picked greedily are maximally different. It is optimized using Approximate Nearest Neighbour algorithm that uses a randomized k-d tree. The max number of comparisons is limited to 100 and the number of trees is limited to 2 to trade off between speed and accuracy. The clustering is done by varying the number of clusters between 1 and 20. In the case where the number of samples is less than the number of clusters, the maximum cluster value is clipped to the maximum number of samples. 
\section{Results}

JANUS CS3 is an extended version of IJBA dataset \cite{IJBA} that contains 11,876 images and 7245 video clips of 1870 subjects. CS3 provides 11 different protocols, that include Identification, Verification and clustering tasks. In our work, we focus on Protocol 6, i.e CS3 1:N Multi Open Set (Video). In Protocol 6 there are 7195 probe templates, where each template is evaluated with respect to two disjoint galleries. There are 940 and 930 templates in Gallery 1 and Gallery 2, respectively. In this case given a video and the annotation of the subject of interest in the first frame, we need to search for a mated template in the gallery for a given probe template. As protocol 6 is an open-set identification problem, there exist some probe templates for which there are no mated templates in the gallery. Therefore, the ranking accuracy is evaluated only for those probe templates that have a mated template in the gallery, demonstrating the closed-set search. For these 20 clusters, the Rank-1, Rank-5, Rank-10, Rank-25, TPIR results are plotted for both JC\cite{JC} and Swami\cite{Swami} features. These results are shown in figure 3.1 to figure 3.6

\begin{figure}[h!]

\centering
\includegraphics[width=0.5\textwidth]{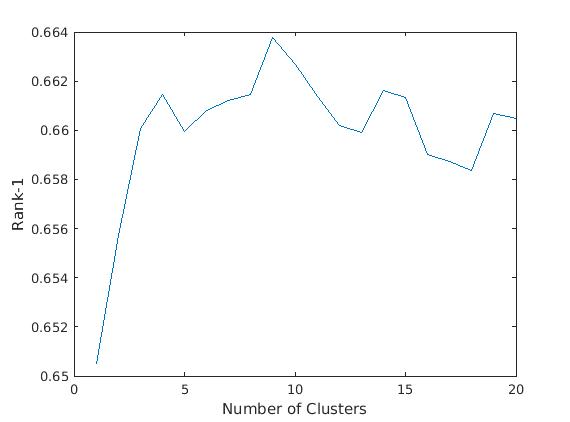}
\label{fig:R1}
\caption{Rank-1 CMC Plot}
\end{figure}

\begin{figure}[h!]

\centering
\includegraphics[width=0.5\textwidth]{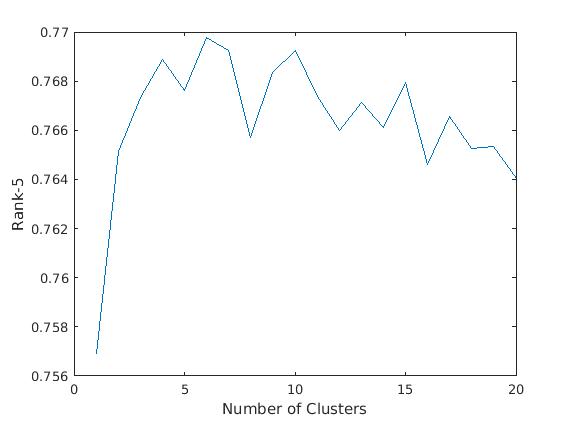}
\label{fig:R5}
\caption{Rank-5 CMC Plot}
\end{figure}

\begin{figure}[h!]

\centering
\includegraphics[width=0.5\textwidth]{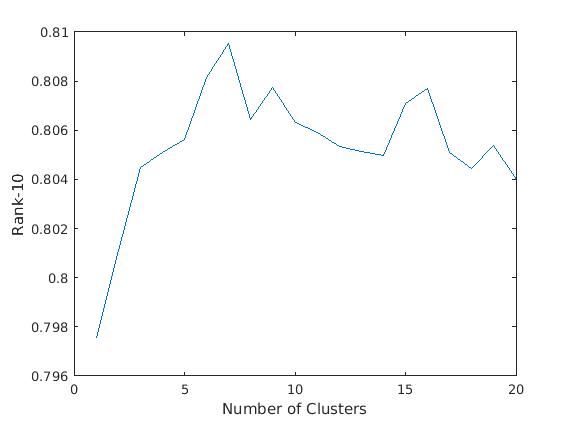}
\label{fig:R10}
\caption{Rank-10 CMC Plot}
\end{figure}

\begin{figure}[h!]

\centering
\includegraphics[width=0.5\textwidth]{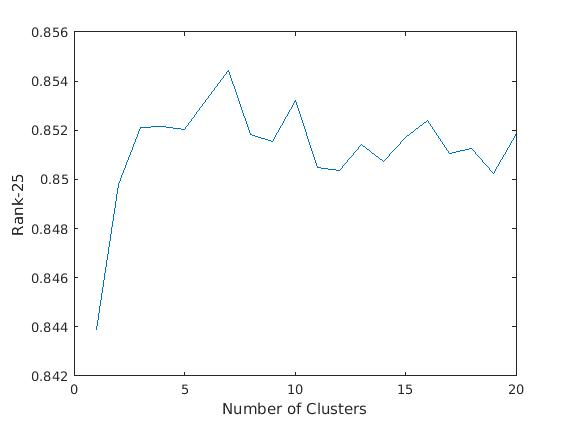}
\label{fig:R25}
\caption{Rank-25 CMC Plot}
\end{figure}

\begin{figure}[h!]

\centering
\includegraphics[width=0.5\textwidth]{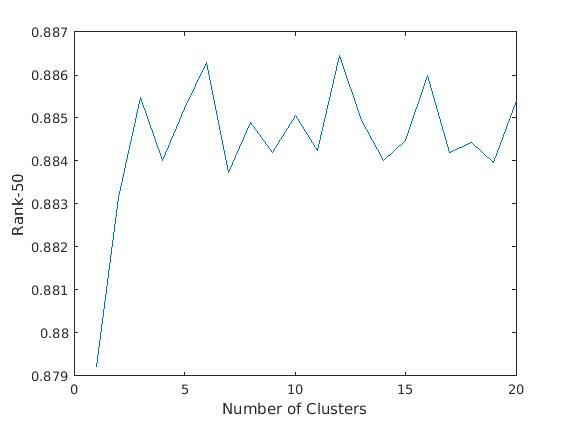}
\label{fig:R50}
\caption{Rank-50 CMC Plot}
\end{figure}

\begin{figure}[h!]

\centering
\includegraphics[width=0.5\textwidth]{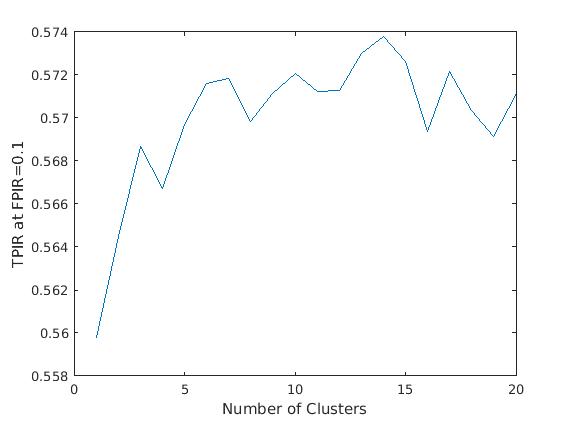}
\label{fig:TPIRFPIR=0.1}
\caption{CMC plot of TPIR at FPIR=0.1}
\end{figure}

\begin{figure}[h!]

\centering
\includegraphics[width=0.5\textwidth]{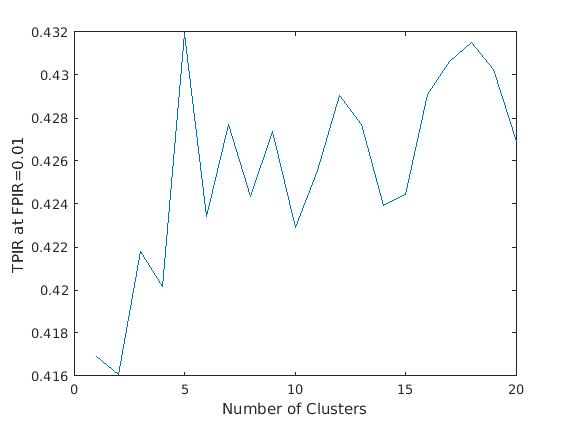}
\label{fig:TPIRFPIR=0.01}
\caption{CMC plot TPIR at FPIR=0.01}
\end{figure}

On an average k=7 clusters work best in respect to Rank-k accuracy and TPIR rate. The computed results for k=7 for JC\cite{JC} are given in table \ref{table:JC_7} and the output on Swami's\cite{Swami} features are given in table \ref{table:Swami_7}. As Ching Hui et al.\cite{TFA} report their results on the average of these two features, we also report the average output in table \ref{table:Average_7}. 

\begin{table}[h!]
\centering

\begin{tabular}{|c|c|c|c|c|c|c|c|}
\hline
          & Rank-1  & Rank-5  & Rank-10 & Rank-25 & Rank-50 & \begin{tabular}[c]{@{}c@{}}TPIR at \\ FPIR=0.1\end{tabular} & \begin{tabular}[c]{@{}c@{}}TPIR at \\ FPIR=0.01\end{tabular} \\ \hline
Gallery 1 & 0.68452 & 0.7913  & 0.83165 & 0.87339 & 0.90226 & 0.59757                                                     & 0.45878                                                      \\ \hline
Gallery 2 & 0.70122 & 0.81287 & 0.84348 & 0.88209 & 0.90678 & 0.61496                                                     & 0.47583                                                      \\ \hline
Average   & 0.72383 & 0.82991 & 0.85983 & 0.89843 & 0.9193  & 0.64765                                                     & 0.50678                                                      \\ \hline
\end{tabular}
\caption{Evaluation of TFA-C using JC features\cite{JC} on k=7 clusters}
\label{table:JC_7}
\end{table}

\begin{table}[h!]
\centering

\begin{tabular}{|c|c|c|c|c|c|c|c|}
\hline
          & Rank-1  & Rank-5  & Rank-10 & Rank-25 & Rank-50 & \begin{tabular}[c]{@{}c@{}}TPIR at \\ FPIR=0.1\end{tabular} & \begin{tabular}[c]{@{}c@{}}TPIR at \\ FPIR=0.01\end{tabular} \\ \hline
Gallery 1 & 0.55    & 0.66782 & 0.71644 & 0.77338 & 0.81806 & 0.46157 & 0.32963                                                      \\ \hline
Gallery 2 & 0.57662 & 0.68472 & 0.7338  & 0.79097 & 0.83449 & 0.46968 & 0.29097                                                    \\ \hline
Average   & 0.59861 & 0.70856 & 0.75926 & 0.81042 & 0.84815 & 0.49606 & 0.34861                                                   \\ \hline
\end{tabular}
\caption{Evaluation of TFA-C using Swami features\cite{Swami} on k=7 clusters}
\label{table:Swami_7}
\end{table}

As we can clearly see, the results for TFA-C in table \ref{table:Average_7} is better than the original TFA algorithm in table \ref{table:TFA} we can state that TFA-C performs better than TFA\cite{TFA}

\begin{table}[h!]
\centering

\begin{tabular}{|c|c|c|c|c|c|c|c|}
\hline
          & Rank-1  & Rank-5  & Rank-10 & Rank-25 & Rank-50 & \begin{tabular}[c]{@{}c@{}}TPIR at \\ FPIR=0.1\end{tabular} & \begin{tabular}[c]{@{}c@{}}TPIR at \\ FPIR=0.01\end{tabular} \\ \hline
Gallery 1 & 0.61726 & 0.72956 & 0.77404 & 0.82339 & 0.86016 & 0.52957 & 0.39421  \\ \hline
Gallery 2 & 0.63892 & 0.7488  & 0.78864 & 0.83653 & 0.87064 & 0.54232 & 0.3834 \\ \hline
Average   &0.66122 & 0.76924 & 0.80954 & 0.85443 & 0.88373 & 0.57186 & 0.4277 \\ \hline
\end{tabular}
\caption{Evaluation of TFA-C using average both features on k=7 clusters}
\label{table:Average_7}
\end{table}

\begin{table}[h!]
\centering

\begin{tabular}{|l|l|l|l|l|l|l|l|}
\hline
          & Rank-1 & Rank-5 & Rank-10 & Rank-25 & Rank-50 & \begin{tabular}[c]{@{}l@{}}TPIR at \\ FPIR=0.1\end{tabular} & \begin{tabular}[c]{@{}l@{}}TPIR at\\ FPIR=0.01\end{tabular} \\ \hline
Gallery 1 & 0.6689 & 0.7875 & 0.8264  & 0.8803  & 0.913   & 0.5701                                                      & 0.3892                                                      \\ \hline
Gallery 2 & 0.5514 & 0.6803 & 0.7315  & 0.7926  & 0.8394  & 0.4245                                                      & 0.2931                                                      \\ \hline
Average   & 0.6101 & 0.7339 & 0.779   & 0.8365  & 0.8762  & 0.4973                                                      & 0.3411                                                      \\ \hline
\end{tabular}
\caption{TFA\cite{TFA} results on average of both features on k=7 clusters}
\label{table:TFA}
\end{table}
\section{Conclusion}
We observe that as the number of clusters(k) are increased for the template aggregation, the identification rate increases to a point and deprecates after that. Based on the averages, we observe cluster numbers, k=7 works the best for identification rate in closed set search as shown by the CMC Rank curves and also in the open set search as shown by the CMC TPIR curves. We conclude that our method TFA-C outperforms the existing TFA algorithm by a significant margin. 
\section{Acknowledgement}
This research is based upon work supported by the Office of the Director
of  National  Intelligence  (ODNI),  Intelligence  Advanced  Research  Projects
Activity  (IARPA),  via  IARPA  R\&D  Contract  No.  2014-14071600012.  The
views and conclusions contained herein are those of the authors and should not
be interpreted as necessarily representing the official policies or endorsements,
either  expressed  or  implied,  of  the  ODNI,  IARPA,  or  the  U.S.  Government.
The  U.S.  Government  is  authorized  to  reproduce  and  distribute  reprints  for
Governmental purposes notwithstanding any copyright annotation thereon

\chapter{Appendix}

\section*{Photo-Sketch}
\begin{figure}[h!]

\centering
\includegraphics[width=0.5\textwidth]{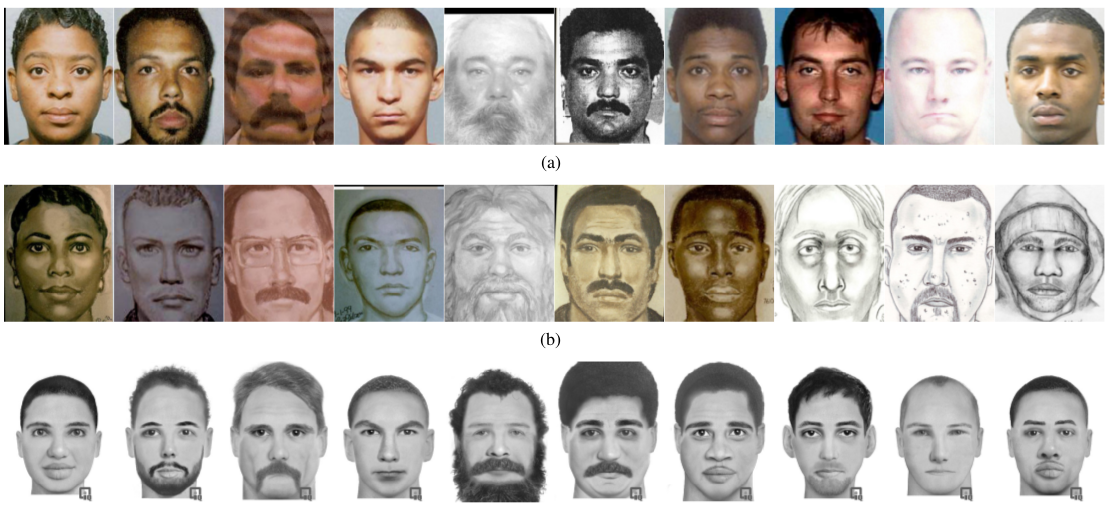}

\caption{Forensic and Composite sketched corresponding to mugshot images \cite{PhotoSketch_Jain}}
\label{fig:FvsC}
\end{figure}

Facial sketches are an essential part of forensics in law enforcement, particularly in those cases where the only evidence is in the form of eye-witness testimony. Facial sketches are of two types, Forensic Sketches that are drawn by forensic artists, and Composite Sketches that are created using computer software \cite{PhotoSketch_Jain}.  Once the sketches are drawn from either of these methods, it allows the law enforcement to apprehend the person of interest based on it. Several works have tried to automate this process by automatically matching\cite{PhotoSketch_Jain} the sketches to the criminal database. Figure \ref{fig:FvsC} shows composite and forensic sketches corresponding to the mugshot images as developed by Klum et al.\cite{PhotoSketch_Jain}. They also show that the matching accuracy of composite sketches is higher than that of the forensic sketches. As evident from the figure \ref{fig:FvsC}, composite sketches are more close to the mugshot images in the domain, and hence they have a better matching accuracy. 

\begin{figure}[h!]

\centering
\includegraphics[width=0.8\textwidth]{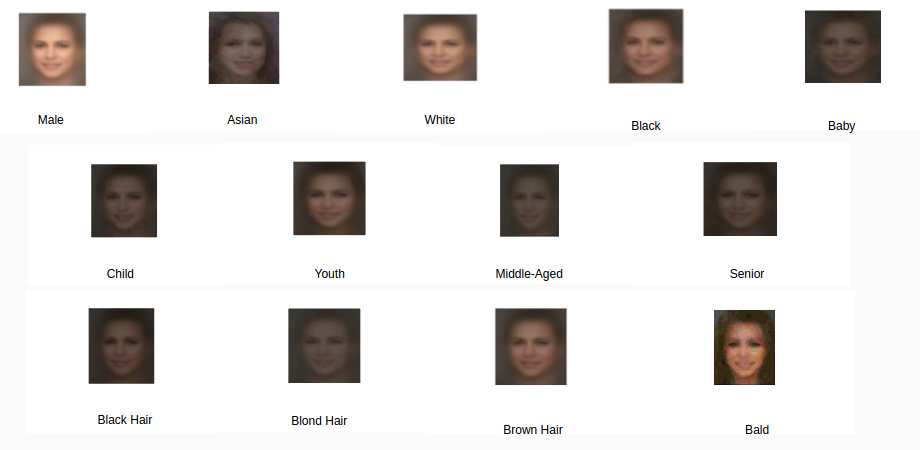}

\caption{Average faces using one attribute}
\label{fig:PS1}
\end{figure}

\begin{figure}[h!]

\centering
\includegraphics[width=0.5\textwidth]{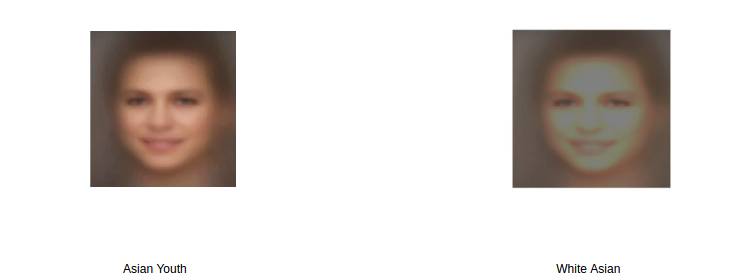}

\caption{Average faces using two attributes}
\label{fig:PS2}
\end{figure}

Motivated by the fact, that at the end the ultimate aim of sketches is matching, we wanted to develop automatic sketches in the mugshot domain. For our work, we used the PubFig dataset to develop single average template faces for the attributes using one attribute and two attributes as shown in figure \ref{fig:PS1} and figure \ref{fig:PS2}

As evident from figure \ref{fig:PS1} and figure \ref{fig:PS2} the average template suffer high illumination artifacts and there is a bias across not only the subjects but across the attributes. So a trade-off needs to met so that the dataset is balanced not only in the subjects but also, attributes. Due to the lack of such a curated dataset and the ill-posed problem, we will like to work further on this problem by either developing a dataset in the future, or utilizing a dataset if any is created that balances classes across not only subjects, but attributes as well.

\begin{vita}
\heading{The author of my thesis} \vspace{15pt}
\begin{descriptionlist}{xxxxx-xxxxx} 
\item[2017] M.S in Computer Science, Rutgers University, USA
\item[2014] B.E in Instrumentation and Control, University of Delhi, India
\end{descriptionlist}
\medskip
\begin{descriptionlist}{xxxxx-xxxxx} 
\item[2016-2017] Graduate assistant, Department of Computer Science, Rutgers University
\item[2015-2016] Teaching assistant, Department of Computer Science, Rutgers University
\item[2015-2016] Grader, Department of Computer Science, Rutgers University
\item[2011-2015] Visiting Researcher, IIT-Delhi, India

\end{descriptionlist}
\end{vita}

\end{document}